\documentclass[letterpaper]{article} % DO NOT CHANGE THIS
\usepackage{aaai23}  % DO NOT CHANGE THIS
\usepackage{times}  % DO NOT CHANGE THIS
\usepackage{helvet}  % DO NOT CHANGE THIS
\usepackage{courier}  % DO NOT CHANGE THIS
\usepackage[hyphens]{url}  % DO NOT CHANGE THIS
\usepackage{graphicx} % DO NOT CHANGE THIS
\usepackage{natbib}  % DO NOT CHANGE THIS AND DO NOT ADD ANY OPTIONS TO IT
\usepackage{caption} % DO NOT CHANGE THIS AND DO NOT ADD ANY OPTIONS TO IT
\usepackage{algorithm}
\usepackage{algorithmic}
\usepackage{amsfonts}
\usepackage{amsmath,amssymb,stmaryrd}
\usepackage{mathtools}
\DeclarePairedDelimiter\floor{\lfloor}{\rfloor}
\usepackage{multirow}
\usepackage{booktabs}
\usepackage{enumitem}
\usepackage{newfloat}
\usepackage{listings}
\DeclareCaptionStyle{ruled}{labelfont=normalfont,labelsep=colon,strut=off} % DO NOT CHANGE THIS
\floatstyle{ruled}
\newfloat{listing}{tb}{lst}{}
\floatname{listing}{Listing}
\title{Let Graph be the Go Board: Gradient-free Node Injection Attack \\ for Graph Neural Networks via Reinforcement Learning}
\author{
Mingxuan Ju\textsuperscript{\rm 1}, Yujie Fan\textsuperscript{\rm 2}, Chuxu Zhang\textsuperscript{\rm 3}, Yanfang Ye\textsuperscript{\rm 1}
}
\affiliations{
    \textsuperscript{\rm 1} University of Notre Dame, Notre Dame, IN 46556\\
    \textsuperscript{\rm 2} Case Western Reserve University, Cleveland, OH 44106\\
    \textsuperscript{\rm 3} Brandeis University, Waltham, MA 02453\\
    {$^1$$\{$mju2, yye7$\}$@nd.edu}; {$^2$yxf370@case.edu}; {$^3$chuxuzhang@brandeis.edu}
}

\begin{document}

\maketitle

\begin{abstract}
Graph Neural Networks (GNNs) have drawn significant attentions over the years and been broadly applied to essential applications requiring solid robustness or vigorous security standards, such as product recommendation and user behavior modeling. 
Under these scenarios, exploiting GNN's vulnerabilities and further downgrading its performance become extremely incentive for adversaries.
Previous attackers mainly focus on structural perturbations or node injections to the existing graphs, guided by gradients from the surrogate models.
Although they deliver promising results, several limitations still exist. 
For the structural perturbation attack, to launch a proposed attack, adversaries need to manipulate the existing graph topology, which is impractical in most circumstances.
Whereas for the node injection attack, though being more practical, current approaches require training surrogate models to simulate a white-box setting, which results in significant performance downgrade when the surrogate architecture diverges from the actual victim model.
To bridge these gaps, in this paper, we study the problem of black-box node injection attack, without training a potentially misleading surrogate model.
Specifically, we model the node injection attack as a Markov decision process and propose \textbf{\underline{G}}radient-free \textbf{\underline{G}}raph \textbf{\underline{A}}dvantage \textbf{\underline{A}}ctor \textbf{\underline{C}}ritic, namely \textbf{G$^2$A2C}, a reinforcement learning framework in the fashion of advantage actor critic. 
By directly querying the victim model, G$^2$A2C learns to inject highly malicious nodes with extremely limited attacking budgets, while maintaining a similar node feature distribution.
Through our comprehensive experiments over eight acknowledged benchmark datasets with different characteristics, we demonstrate the superior performance of our proposed G$^2$A2C over the existing state-of-the-art attackers. Source code is publicly available at: \url{https://github.com/jumxglhf/G2A2C}.
\end{abstract}

\section{Introduction}

Graph neural networks (GNNs), a class of deep learning methods designed to perform inference on graph data, have achieved outstanding performance in various real-world applications, such as recommendation system \cite{ying2018graph}, user behavior modeling \cite{pal2020pinnersage} and drug discovery \cite{jiang2021could}.
The success of GNNs relies on their powerful capability of integrating the graph structure and node features simultaneously for node representation learning. Specifically, the majority of popular GNNs \cite{kipf2016semi,velivckovic2017graph} follow a neural message-passing scheme to learn node embeddings via recursively aggregating and propagating neighbor information.
Along with their great success, the robustness of GNNs has also attracted increasing attentions in recent years, and it has been proved that such a message-passing scheme is vulnerable to adversarial attacks \cite{zugner2018adversarial,chen2021understanding}. 
% Particularly, generating unnoticeable perturbations on graph structure or node features are able to degrade the learned node representations, and consequently deteriorate the performance of GNNs in downstream tasks. 

Existing research efforts on graph adversarial attack mainly concentrate on graph structure perturbations via modifying edges \cite{dai2018adversarial,wang2019attacking,zugner2018adversarial}. 
Despite their promising performance, these attack strategies have narrow applications since the adversaries are required to manipulate the existing graph topology, which is impractical under most circumstances.
Besides graph structural perturbations, another trend of research focuses on the node injection attacks \cite{tao2021single,zou2021tdgia,wang2020scalable}.
They explore a more practical setting where attacks are launched by injecting new nodes into the existing graphs, and hence the authorities of modifying the existing graph structures are unnecessary. 
Considering spam detection in the social networks as an example, where adversaries aim at tricking the victim model into misclassifying the spam accounts (i.e., target nodes).
In many circumstances, they do not have permission to add or remove the friendships already formed among the existing users (i.e., modifying connections between existing nodes).
However, adversaries can easily create accounts with new profiles and establish new links with the existing users to fulfill the attack purposes (i.e., injecting new malicious nodes to deceive the victim model).
Apparently, the node injection attacks are more feasible compared with the attacks via graph perturbations.

Nevertheless, node injection attacks are challenging, and the adversaries should consider: (i) \textit{how to generate imperceptible yet malicious features for the injected nodes?} and (ii) \textit{how to establish links between an injected node and the existing nodes?}
Current related works \cite{tao2021single,zou2021tdgia,wang2020scalable,sun2019node} fulfill these purposes according to the gradient from the victim model in the white-box setting or the simulated gradient from the surrogate model in the black-box setting. 
Though promising, white-box approaches are usually not practical, and the performance of black-box ones can be deteriorated when the surrogate architecture and the actual victim model diverge.

In this work, we consider the most challenging and practical scenario, i.e., black-box evasion attack through the node injection, where only adjacency matrix, node attributes and model queries are available. 
To tackle these aforementioned challenges, we propose \textbf{\underline{G}}radient-free \textbf{\underline{G}}raph \textbf{\underline{A}}dvantage \textbf{\underline{A}}ctor \textbf{\underline{C}}ritic, namely \textbf{G$^2$A2C}, a reinforcement learning framework in the fashion of advantage actor critic. 
Different from the existing counterparts, G$^2$A2C does not require the adversaries to train a surrogates model since the vulnerabilities of the victim model are learned according to the queries from the victim model instead of the surrogate gradient.
Thus, G$^2$A2C makes no prior assumption about the victim model, which eliminates possible performance downgrade introduced by the divergence between the assumption and the actual victim model.
Specifically, we formulate the node injection attack as a Markov Decision Process (MDP), where the attack is decoupled into node generation and edge wiring. 
During the node generation phase, imperceptible yet malicious features are attributed to the adversarial node. 
To guarantee the imperceptibility, besides the regularization on the similarities to the real nodes from the original graph, we design separate strategies to tackle both discrete and continuous feature spaces, so that G$^2$A2C can inject nodes according to the real feature space. 
And during the edge wiring phase, edges between the injected node and the remaining graph are wired according to a learnable conditional probability distribution. 
These two steps are gradient-free because they are guided by the rewards calculated from the model feedback instead of the surrogate gradients. 
The key contributions of this paper are summarized as follows:
\begin{itemize}[leftmargin=*]
    \item This is the first work that studies black-box node injection attack for GNNs without using the surrogate gradient, eliminating the performance downgrade entailed by the inaccurate approximation of the victim model.
    \item We carefully formulate the black-box node injection attack as an MDP and propose G$^2$A2C to launch effective yet imperceptible attacks against GNNs trained on graphs with either discrete or continuous node features.
    \item With comprehensive experiments over eight acknowledged benchmark datasets, we demonstrate G$^2$A2C's superior attack effectiveness with different attack budgets by comparison with the state-of-the-art attack models. 
\end{itemize}

\begin{figure*}[!h]
\centering
\includegraphics[width=1\linewidth]{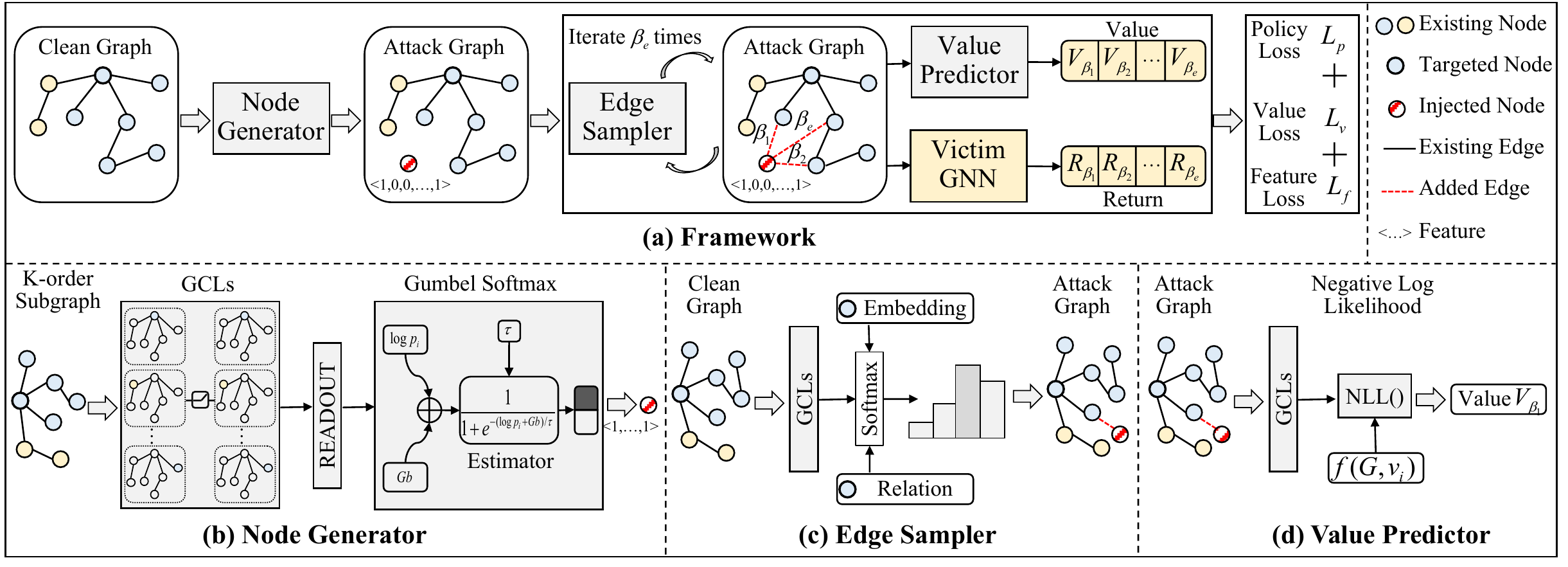}
\caption{System overview of G$^2$A2C.
% . In order for the victim GNN to misclassify the target node (i.e., the node with bold circle), the node generator attributes the injected node (i.e, the node with red dashed line) with imperceptible yet malicious features, considering the target nodes' sub-graph, as shown in (b). Given the generated node, the edge sampler wires it to the existing graph according to a learnable conditional distribution and the attack budget, as shown in (c). This two modules are guided by the value predictor, which predicts the expected future value, supervised by queries from the victim model, as shown in (d). This process iterates until the target node is successfully attacked or the attack budget is depleted.
}
% \vspace{-0.4cm}
\label{fig:system}
\end{figure*}

\section{Preliminary}

Let $G = (V,E)$ denote a graph, where $V$ is the set of $|V| = N$ nodes and $E \subseteq V \times V$ is the set of $|E|$ edges between nodes. 
Adjacency matrix is denoted as $\textbf{A} \subseteq \{0,1\}^{N \times N}$, where $a_{ij}$ at $i$-th row and $j$-th column equals to 1 if there exists an edge between nodes $v_i$ and $v_j$, and 0 otherwise.
We further denote the node feature matrix as $\mathbf{X} \in \mathbb{R}^{N \times F}$ where node $v_i$ is associated with a feature vector $\mathbf{x}_i \in \mathbb{R}^F$ of dimension $F$. 
$\mathbf{Y} \subseteq \{0,1\}^{N \times C}$ denotes the label matrix of a graph, where $C$ is the number of total classes. 
For $M$ labeled nodes ($0 < M \ll N$) with label $\mathbf{Y}^L$ and $N-M$ unlabeled nodes with missing label $\mathbf{Y}^U$, the objective of GNNs for node classification is to predict $\mathbf{Y}^U$ given $\mathbf{Y}^L$, $\mathbf{A}$ and $\mathbf{X}$. 

\subsection{Graph Neural Network}

GNNs generalize neural networks into graph-structured data \cite{kipf2016semi,velivckovic2017graph,klicpera2019predict,ju2022adaptive}. The key operation is graph convolution where information is routed between nodes with some pre-defined deterministic rules (e.g., adjacency matrices, Laplacian matrices, and attention). 
For example, the graph convolution layer (GCL) of GCN \cite{kipf2016semi} is formulated as: $\mathbf{H}^{(l+1)} = f_{GCL}^{(l)}(\hat{\mathbf{A}}, \mathbf{H}^{(l)},\mathbf{W}^{(l)})= \sigma(\hat{\mathbf{A}} \mathbf{H}^{(l)} \mathbf{W}^{(l)})$,
where $\hat{\mathbf{A}}$ denotes the normalized adjacency matrix with self-loop, $\sigma(.)$ denotes the non-linearity function, and $\mathbf{W}^{(l)}$ and $\mathbf{H}^{(l)}$ are the learnable parameters and node representations at $l^{th}$ layer respectively. 
Normally, at $K$-th layer, with the last dimension of $\mathbf{W}^{(K)}$ and $\sigma(.)$ set to $C$ and softmax respectively, the loss for node $v_i$ is calculated as:
\begin{equation}
	\mathcal{L}(v_i, G, \mathbf{y}_i) = CE\big( f_{GCL}^{(K)}(\hat{\mathbf{A}}, \mathbf{H}^{(K)},\mathbf{W}^{(K)})[i], \mathbf{y}_i\big),
\end{equation} 
where $[\cdot]$ is the indexing operation and $CE(\cdot,\cdot)$ refers to the cross entropy loss function.

\subsection{Graph Adversarial Attack}
For a trained GNN model $f(\cdot,\cdot): V \times G \rightarrow \mathbf{Y}$, the attacker $g(\cdot,\cdot): G \times f \rightarrow G$ is asked to modify the graph $G=(V,E)$ into $G'=(V', E')$ such that:
\begin{equation}
	\begin{split}
		\max_{G'} \;\;\;\;& \mathbb{I}(f(V^U, G') \neq \mathbf{Y}^U) \\
		s.t. \;\;\;\; & G' = g(G,f) \text{\;\;and\;\;} \mathcal{I}(G, G') = 1.
	\end{split}
\label{eq:constraint}
\end{equation}
Here $V^U$ can be the testing set or nodes of interest, $\mathbb{I}(\cdot)$ is an indicator function that returns the number of true conditions, and $\mathcal{I}(\cdot, \cdot):G\times G \rightarrow \{0,1\}$ is an indicator function, which returns one if two graphs are equivalent under the classification semantics. 
There exist two approaches to fulfill the attack purposes. The first is edge modification in $G$, also known as structural perturbation, which changes the entries in $\mathbf{A}$. Whereas the second approach tampers the nodes via adding, modifying, or deleting nodes in $G$, resulting in not only entry-level but also dimensional changes to both $\mathbf{A}$ and $\mathbf{X}$. In this work, we focus on the black-box node injection evasion attack, a special case of the second approach, where three attack budgets need to be considered: the number of adversarial nodes injected per attacking one node, denoted as $\beta_n$, the degree of each injected node $\beta_e$, and lastly the feature distribution shift $\beta_f$. Hence, $\mathcal{I}(\cdot, \cdot)$ is defined as
% Hence, the indicator function $\mathcal{I}(\cdot, \cdot)$ in Eq. (\ref{eq:constraint}) is defined as
\begin{equation}
\begin{split}
    \mathcal{I}&(G, G') = \mathbb{I}(|V'| - |V| \leq \beta_n) \cdot \prod_{i=N}^{N+\beta_n}\mathbb{I}(\rVert \mathbf{a}_i'\rVert \leq \beta_e) \\
    & \cdot  \prod_{i=N}^{N+\beta_n}\mathbb{I}(\langle \mathbf{x}'_i, \mathbf{X} \rangle \leq \beta_f),
\end{split}
\end{equation}
where $\langle\cdot,\cdot\rangle$ refers to the metric measuring the similarities between the generated feature and the original features (e.g., Kullback–Leibler divergence for continuous features, and norm difference for discrete features).

\section{Methodology}
Given a clean graph $G = (V,E)$, the attacker $g$ injects a set of adversarial nodes $V^{\mathcal{A}}$ with generated features $\mathbf{X}^{\mathcal{A}}$ into the clean node set $V$. 
After injecting $V^{\mathcal{A}}$, attacker $g$ creates adversarial edges $E^{\mathcal{A}} \subseteq V^{\mathcal{A}} \times V \cup V^{\mathcal{A}} \times V^{\mathcal{A}}$ to evade the detection of GNN $f$ for nodes $V^U$. 
$G' = (V', E')$ is the attacked graph in which $V' = V \cup V^{\mathcal{A}}$, $E' = E \cup E^{\mathcal{A}}$, and $\mathbf{X}' = \mathbf{X} \oplus \mathbf{X}^{\mathcal{A}}$, where $\oplus$ is the vertical concatenation.

Injecting node involves generating discrete graph data, such as adjacency matrices or feature matrices, that gradient-based approaches handle poorly in many circumstances. 
This phenomenon could be further aggravated by the black-box setting where gradient information from the surrogate model might not be accurate.
Moreover, generating node and assigning edges are naturally sequential and reinforcement learning fits for such Markov Decision Process (MDP).
Hence, to perform the optimization task in Eq. (\ref{eq:constraint}), we propose to explore deep reinforcement learning.
Specifically, we utilize an on-policy A2C reinforcement learning framework, adapted from \cite{mnih2016asynchronous}, instead of the off-policy algorithms such as deep Q-learning. Since A2C circumvents the need to calculate the expected value for every possible action, which is intractable. % (e.g., Q learning computes $F \choose \rVert\mathbf{x}\rVert$ Q values, exceeding $10^{41}$ possibilities with a common selection of $F=1000$ and $\rVert\mathbf{x}\rVert=20$; whereas on-policy A2C requires only one step). 

The overview of G$^2$A2C is shown in Figure \ref{fig:system}.
Given a graph $G$ and a target node $v_i$, the node generator $g_n$ creates the adversarial node according to $v_i$'s sub-graph. 
Then edge sampler $g_e$ is forwarded for $\beta_e$ consecutive times to connect the injected node to the existing graph. Previous two processes iterate until the label of $v_i$ has been successfully changed or the attack budget is depleted. A detailed definition of our proposed MDP is defined as follows:

\paragraph{State.} $s_t \in S$ contains the intermediate modified graph $G'_t = (V'_t, E'_t)$ as well as the generated features $\mathbf{X}^{\mathcal{A}}_t$ at the timestamp $t$. 
To efficiently interpret $s_t$, edge sampler $g_e$ and node generator $g_n$ attend to the $K$-hop sub-graph $G'_t(v_i)$ entailed by the target node $v_i$, where $K$ is a hyper-parameter for the number of stacked GCLs. 
We restrain our scope on the neighbors within $K$ hops since normally the victim GNN $f$ has a shallow receptive field. 
However, without loss of generality, $g_e$ can be extended to the full graph $G'_t$ for the optimal performance.

\paragraph{Action.} Node injection attack can be decoupled into two components: creating the injected node and wiring it to the existing graph. 
We model this process as an MDP where G$^2$A2C starts with the node generation and then wires the generated node to the existing graph for $\beta_e$ consecutive times. 
The MDP terminates if the attacker $g$ successfully evades the detection from $f$ or the attack budget is depleted. 
Formally, at time $t$, the node generation action is denoted as $a^{(n)}_t$, and the node wiring action is denoted as $a^{(e)}_t$.
The trajectory of our proposed MDP is ($s_0$, $a^{(n)}_0$, $s_1$, $a^{(e)}_1$, $r_1$, $s_2$, $a^{(e)}_2$, $r_2$, $s_2$, $\dots$,$a^{(n)}_t$, $s_t$, $\dots$, $s_T$), where $s_T$ refers to the terminal state, and $r_t$ is the reward for the action $a_t$. 

\paragraph{Reward.} In our curated setup, generating an isolated node does not entail a reward value, as an isolated node brings no perturbation to $f$; instead, the impact is reflected later when links are wired to the existing graph.
Hence, reward values are only assigned to the edge wiring actions. 
During the intermediate phase of attacking node $v_i$ at timestamp $t$, the reward for edge wiring action $a^{(e)}_t$ is calculated as:
\begin{equation}
\begin{split}
  r(v_i, a^{(e)}_t, G'_t) &= \mathcal{L}\big(v_i, G'_{t+1}, f(v_i, G)\big)\\
  &- \mathcal{L}\big(v_i, G'_t, f(v_i, G)\big) \;\; \text{if}\; s_{t+1} \neq s_T,
\end{split}
  \label{eq:reward1}
\end{equation}
where $G'_{t+1}$ is the resulted graph after applying $a^{(e)}_t$ to $G'_t$. 
The reward function measures the difference between the classification losses before and after the edge wiring, which encourages G$^2$A2C to imperil the correct decision of the victim model. 
Besides, to further motivate our model to actively evade the detection, we give extra rewards if the prediction of $v_i$ is flipped at the end of one attacking episode (i.e., $\mathbb{I}\big(f(G'_T, v_i) \neq f(v_i, G)\big)$).

\subsection{Node Injection Attack via Actor Critic}

\paragraph{Adversarial Node Generator}

To deceive $f$ into misclassifying the target node $v_i$, given its K-order sub-graph $G'_t(v_i)$, the node generator $g_n$ aims to create an adversarial node $v_a$ with an imperceptible yet malicious feature vector $\mathbf{x}_a$.
Specifically, the generated $\mathbf{x}_a$ should follow the same characteristic conventions as $\mathbf{X}$. We should not expect a continuous $\mathbf{x}_a$ when all other feature vectors are discrete, and vice versa. 
% In this paper, we study the special and far less researched discrete feature generation, since binary feature vectors are commonly seen in the real world. 
Moreover, $\mathbf{x}_a$ should not diverge too much from nodes features in $G'_t(v_i)$, as restrained by the distribution shift budget $\beta_f$. 
To tackle the aforementioned challenges, $g_n$ is equipped with $K$-stacked GCLs, conducts message propagation on $G'_t(v_i)$, summarizes $G'_t(v_i)$ by a readout function \cite{xu2018powerful}, and with Gumbel-Softmax \cite{jang2016categorical}, generates $\mathbf{x}_a$ tailoring the vulnerability of $v_i$ as well as the imperceptibility, as shown in Figure \ref{fig:system} (b).
Formally, the $K$-th convolution layer of $g_n$ can be described as: 
\begin{equation}
        \mathbf{H}^{(K+1)}_n = f_{GCL}^{(n,K)}\big(\hat{\mathbf{A}}(v_i), \mathbf{H}^{(K)}_n,\mathbf{W}^{(K)}_n\big),
    \label{eq:gc}
\end{equation}
where $\mathbf{H}^{(0)}_n=\mathbf{X}'_t$, $\hat{\mathbf{A}}(v_i)$ refers the normalized adjacency matrix of $G'_t(v_i)$ and $\mathbf{W}^{(K)}_n$ is the parameter matrix of $g_n$'s $K$-th convolution layer. 
To consider the holistic representation of $G'_t(v_i)$ and the unique characteristics tailored by $v_i$, we formulate the feature distribution $\mathbf{z}_n$ as:
\begin{equation*}
    \mathbf{z}_n = \sigma\Big(\big(\mathrm{READOUT}(\mathbf{H}^{(K+1)}_n)||\mathbf{H}^{(K+1)}_n(v_i)\big) \cdot \mathbf{W}_n^f\Big),
\end{equation*}
where $\mathrm{READOUT}(\cdot)$ refers to the graph pooling function such as column-wise summation \cite{xu2018powerful}, $\mathbf{H}^{(K+1)}_n(v_i)$ denotes $v_i$'s node embedding after the propagation, and $\mathbf{W}_n^f \in \mathbb{R}^{2d \times F}$ is the learnable parameter matrix that combines the target node's characteristics with the local neighborhood information. 

To inject nodes in the discrete feature space, we directly utilize $\mathbf{z}_n$ as the logits of a relaxed Bernoulli probability distribution, a binary special case of the Gumbel-Softmax reparameterization trick which is soft and differentiable \cite{jang2016categorical}. 
Utilizing the relaxed sample, we apply a straight-through gradient estimator \cite{bengio2013estimating} that rounds the relaxed sample in the forward phase. 
In the backward propagation, actual gradients are directly passed to the relaxed samples instead of the previously rounded values, making $g_n$ trainable. Formally, the discrete version of $\mathbf{x}_a$ is generated by:
\begin{equation}
    \mathbf{x}_a[i] = \floor*{\frac{1}{1+e^{-(\log \mathbf{z}_n[i] + Gb)/\tau}} + \frac{1}{2}},
\label{eq:gumbel}
\end{equation}
where $[\cdot]$ is the indexing operation, $Gb \sim Gumbel(0,1)$ is a Gumbel random variable and $\tau$ is the temperature for the Gumbel-Softmax distribution. 
To maintain the imperceptibility of $\mathbf{x}_a$, we propose a feature loss function for the discrete feature space:
\begin{equation}
    \mathcal{L}_f(\mathbf{x}_a) =\big(\rVert\mathbf{x}_a\rVert/\rVert\mathbf{X}\rVert - \beta_f\big)^2.
\end{equation}
Whereas to inject nodes in the continuous feature space, we transform $\mathbf{z}_n$ into $\mathbf{\mu}_n$ and $\mathbf{\sigma}_n$ by two learnable weight matrices $\mathbf{W}_\sigma$ and $\mathbf{W}_\mu$ $\in \mathbb{R}^{d\times d}$, and utilize them as parameters of a normal distribution to generate the feature, formulated as:
\begin{equation}
    \begin{split}
        & \;\;\;\;\;\;\;\;\;\;\;\;\;\;\;\;\;\; \mathbf{x}_a \sim \mathcal{N}(\mathbf{\mu}_n, \mathbf{\sigma}_n^2) \\
        & \text{where \;\;} \mathbf{\mu}_n=\mathbf{z}_n \cdot \mathbf{W}_\mu \text{\;and\;} \mathbf{\sigma}_n=\mathbf{z}_n \cdot \mathbf{W}_\sigma.
    \end{split}
\end{equation}
For nodes in the continuous feature space, we enforce the imperceptibility by minimizing the KL-divergence between the feature generation distribution and the real node feature distribution (i.e., represented by the mean value of the real features $\mathbf{\mu}_\mathbf{x}$ and their standard deviation $\mathbf{\sigma}_\mathbf{x}$), as follows:
\begin{equation}
\begin{split}
    &\mathcal{L}_f(\mathbf{x}_a) = p(\mathbf{x}_a, \mathbf{\mu}_n, \mathbf{\sigma}_n) \log\frac{p(\mathbf{x}_a, \mathbf{\mu}_n, \mathbf{\sigma}_n)}{p(\mathbf{x}_a, \mathbf{\mu}_\mathbf{x}, \mathbf{\sigma}_\mathbf{x})} - \beta_f, \\
     & \;\;\;\;\;\;\text{where \;\;} p(\mathbf{x}_a, \mu, \sigma)= \frac{1}{\sigma \sqrt{2\pi}} e^{-\frac{1}{2}(\frac{(\mathbf{x}_a)-\mu}{\sigma})^2} .
\end{split}
\end{equation}

\paragraph{Adversarial Edge Sampler}
For the target node $v_i$, given the generated node features $\mathbf{x}_a$ from $g_n$, the adversarial edge sampler $g_e$ aims at connecting $v_a$ to the current graph $G'_t$. 
Similar to $g_n$, $g_e$ is equipped with a $K$-stacked GCLs, formulated as $\mathbf{H}^{(K+1)}_e = f_{GCL}^{(e,K)}(\hat{\mathbf{A}}(v_i), \mathbf{H}^{(K)}_e,\mathbf{W}^{(K)}_e)$, where $\mathbf{H}^{(0)}_e = \mathbf{X}'$, and $\mathbf{W}^{(K)}_e$ is the parameter matrix of $g_e$'s $K$-th convolution layer. 
Then, we concatenate $\mathbf{x}_a$ with each row of $\mathbf{H}^{(K+1)}_e$ to obtain $\mathbf{Z}_e \in \mathbb{R}^{|V'| \times (d+F)}$.
The probability vector of remaining nodes connecting to $v_a$ is calculated as:
\begin{equation}
    \mathbf{p}_e = \text{softmax}(\mathbf{Z}_e \cdot \mathbf{W}_e + \mathbf{A}[v_i]),
    \label{eq:edge_prob}
\end{equation}
where $\mathbf{W}_e \in \mathbb{R}^{(d+F)}$ is the learnable parameter matrix and $\mathbf{A}[v_i]$ denotes $v_i$'s row in the adjacency matrix of $G'_t$.
We add $\mathbf{A}[v_i]$ to the probability logits because in order for the adversarial perturbation to be perceived by $f$, the introduced edge must enable $\mathbf{v}_a$ to lie in the receptive field of $f$. 
% Manually increasing the probabilities of $v_a$ connecting to the first-order neighbors of $v_i$ speeds up the convergence time.
Then, we sample an edge from an one-hot categorical distribution parameterized by $\mathbf{p}_e$, merge the sampled edge into $G'_t$ and get $G'_{t+1}$.
For the next edge sampling operation, $G'_{t+1}$ is fed into $g_e$, and this process iterates until $v_i$ is successfully evaded or the number of wired edges reaches to $\beta_e$, as shown in Figure \ref{fig:system} (c).

\paragraph{Value Predictor}

Along with the policy learners $g_n$ and $g_e$ we have proposed, the value predictor $g_v$ is the other component of A2C that aims at predicting the expected accumulated rewards at the end of the MDP. 
Given the dedicated reward function we have defined in Eq. (\ref{eq:reward1}), $g_v$ predicts the final accumulated loss score of targeted node based on the current $G'_t$.
We formulate this process as a regression task, where $g_v$ predicts the negative log likelihood between the class log probabilities in current graph $G'_t$ and $f(G, v_i)$. 
Specifically, a GNN model with $K$-stacked layers is utilized to capture the node topological information, similar to $g_e$, formulated as: $\mathbf{H}^{(K+1)}_v = f_{GCL}^{(v,K)}(\hat{\mathbf{A}}, \mathbf{H}^{(K)}_v,\mathbf{W}^{(K)}_v)$, where $\mathbf{W}^{(K)}_v$ is the parameter matrix of $g_v$'s K-th convolution layer. As shown in Figure \ref{fig:system} (d), we extract node $v_i$'s embedding and concatenate it with $f$'s output to predict the value score, formulated as:
\begin{equation}
\begin{split}
     & g_v(v_i, G'_t) = \\
     & NLL\Big(\big(\mathbf{H}^{(K+1)}_v(v_i)||f(v_i, G'_t)\big)\cdot \mathbf{W}_v, f(v_i, G)\Big)
\end{split}
\end{equation}
where $NLL(\cdot, \cdot)$ is the negative log likelihood function, and $\mathbf{W}_v \in \mathbb{R}^{(d+C) \times C}$ is the learnable parameter.
\subsection{Training Algorithm}
To train G$^2$A2C = $\{g_n, g_e, g_v\}$, we explore the experience replay technique with memory buffer $\mathcal{M}$. 
Intuitively, we simulate the selection process to generate the training data and store the experience in the memory buffer during the forward runs of training phase. 
An instance in $\mathcal{M}$ is in the format of triplet $(G'_t, a_t, R_t)$ with return $R_t = \sum_{j=t}^{j=T} r(v_i, a_j, G'_j) \cdot \gamma^{(j-t)}$, where $\gamma$ refers to the discount factor. 
During the back-propagation, three losses are involved: policy loss $\mathcal{L}_p$, value loss $\mathcal{L}_v$ and feature loss $\mathcal{L}_f$. 
Given a triplet $(G'_t, a_t, R_t) \in \mathcal{M}$, $\mathcal{L}_p$ is calculated as:
\begin{equation*}
     \mathcal{L}_p(G'_t, a_t, R_t) = - \log\big(p(a_t|G'_t)\big) \cdot \big(R_t - g_v(v_i, G'_t)\big),
\end{equation*}
where $p(a_t|G'_t)$ denotes the probability of conducting action $a_t$ under the graph $G'_t$. In $\mathcal{L}_p$, the second term $(R_t - g_v(v_i, G'_t))$ is also known as the advantage score \cite{mnih2016asynchronous}, which depicts how much better of selecting action $a_t$ over the other actions. $\mathcal{L}_p$ enforces G$^2$A2C to deliver better actions with higher probabilities. On the other hand, value loss $\mathcal{L}_v$ enforces the value predictor $g_v$ to correctly deliver the actual accumulated reward, calculated as: 
\begin{equation}
     \mathcal{L}_v(G'_t, R_t) = |g_v(v_i, G'_t) - R_t|.
\end{equation}
The final loss for G$^2$A2C is formulated as:
\begin{equation*}
\mathcal{L} = \sum_{\mathcal{M}}\big( \mathcal{L}_v(G'_t, R_t) + \mathcal{L}_p(G'_t, a_t, R_t)\big) + \sum_{\mathbf{x}_a \in \mathbf{X}_t^{\mathcal{A}}} \mathcal{L}_f(\mathbf{x}_a).
\end{equation*}

\section{Experiment}
In this section, we aim at answering the following four research questions: (\textbf{RQ1}) Can our proposed G$^2$A2C effectively evade target nodes given a well trained GNN for various datasets? (\textbf{RQ2}) Can the ``gradient-free'' property enhance attack performance when inaccurate victim model architecture is approximated? (\textbf{RQ3}) What is the attack performance of G$^2$A2C under different budgets? (\textbf{RQ4}) How does G$^2$A2C conduct the node injection attack in real cases? 

\paragraph{Dataset.} We conduct experiments on eight acknowledged benchmark datasets, namely \texttt{Cora}, \texttt{Citeseer}, \texttt{Pubmed} \cite{sen2008collective}, \texttt{Amazon Photo}, \texttt{Amazon Computers} \cite{mcauley2015inferring}, \texttt{Wiki. CS} \cite{mernyei2020wiki}, \texttt{Reddit} \cite{hamilton2017inductive}, and \texttt{OGB-Products} \cite{hu2020open}. 
These datasets cover a broad range of fields, such as social networks, merchandise networks and citation networks. 
Besides, node features of these datasets cover both discrete and continuous spaces, to validate the attack performance of all baselines under various scenarios. 
We use public splits for training and evaluation on \texttt{Cora}, \texttt{Citeseer} and \texttt{Pubmed}, the random splits of 10\%/10\%/80\% for \texttt{Amazon Photo}, \texttt{Amazon Computers} and \texttt{Wiki. CS}. 
For \texttt{OGB-Products} and \texttt{Reddit}, we explore the sub-graphs and splits shared by G-NIA \cite{tao2021single} for fair comparison. 
The dataset statistics are shown in Table \ref{dataset}.
% \vspace{-0.1cm}
\paragraph{Baselines.} Since black-box node injection attack is an emerging and far less researched area, only few methods focus on this topic, such as NIPA \cite{sun2019node}, G-NIA \cite{tao2021single}, AFGSM \cite{wang2020scalable}, and TDGIA \cite{zou2021tdgia}. 
To sufficiently demonstrate the effectiveness of G$^2$A2C, besides these methods, we also compare G$^2$A2C with the adaptions of the state-of-the-art structural perturbation method Nettack \cite{zugner2018adversarial}. Accordingly, our baselines include: NIPA \cite{sun2019node} and its variant (i.e., Node+NIPA), two variants of Nettack \cite{zugner2018adversarial} (i.e., Rand.+Nettack and Node+Nettack), G-NIA \cite{tao2021single}, AFGSM \cite{wang2020scalable}, and TDGIA \cite{zou2021tdgia}. 
``Rand'' and ``Node'' refer to the random-generated and the G$^2$A2C-generated node features, respectively.  
To compare G$^2$A2C with white-box (i.e., Nettack, AFGSM and G-NIA) or grey-box (i.e., NIPA) approaches requiring gradient from the victim model, we train a 2-layer GCN as the surrogate model.
% We list the detailed explanations of all the baseline models in the technical appendix.

% and structural perturbations are then explored to wire them to the existing graph.
\begin{table}
    \centering
    \begin{tabular}{l|cccc}
    \toprule
    Dataset & Node & Edge & Class & Dim.\\
    \midrule
    \multicolumn{5}{c}{Datasets with Discrete Feature Space} \\
    \midrule
    Cora$^\ast$ & 2,708 & 5,429 & 7 & 1,433  \\
    Citeseer$^\ast$& 3,327 & 4,732 & 6 & 3,703  \\
    Am. Photo   & 7,650 &119,043& 8 & 745    \\
    Am. Comp.& 13,752&245,778& 10& 767    \\
    \midrule
    \multicolumn{5}{c}{Datasets with Continuous Feature Space} \\
    \midrule
    Pubmed$^\ast$  & 19,717 & 44,338  & 3  & 500 \\
    Wiki. CS    & 11,701 & 216,123 & 10 & 300  \\
    OGB-Prod.$^\star$   & 10,494 & 77,656  & 35 & 100  \\
    Reddit$^\star$      & 10,004 & 37,014  & 41 & 602 \\
    \bottomrule
    \end{tabular}
    % \vspace{-0.2cm}
    \caption{Dataset statistics. For the datasets with $\ast$, we explore the public splits \cite{kipf2016semi}. And for datasets with $\star$, we use the largest connect component sub-graphs acquired from \citet{tao2021single}.}
    % \vspace{-0.3cm}
    \label{dataset}
\end{table}
% \vspace{-0.1cm}
\begin{table*}
\centering
\begin{tabular}{l|cccc|cccc}
    \toprule
    \toprule
    \multirow{2}{*}{Attacker} & \multicolumn{4}{c|}{Discrete Feature Space} & \multicolumn{4}{c}{Continuous Feature Space}\\
    \cmidrule(r){2-5} \cmidrule(r){5-9} 
    &               Cora               & Citeseer         & Am. Comp.        & Am. Photo       & OGB-Prod.        & Reddit           & Pubmed           & Wiki. CS\\
    \midrule
    Clean         & 18.4               & 21.1             & 24.37            & 17.8             & 24.3             & 8.5              & 21.9             & 21.3      \\
    NIPA          & 18.6$_{\pm 0.1}$   & 21.1$_{\pm 0.}$  & 25.0$_{\pm 0.2}$ & 17.8$_{\pm 0.}$  & 25.9$_{\pm 0.2}$ & 12.5$_{\pm 0.7}$ & 21.9$_{\pm 0.}$  & 25.2$_{\pm 0.4}$ \\
    Node+NIPA     & 25.3$_{\pm 0.4}$   & 33.5$_{\pm 0.6}$ & 32.6$_{\pm 0.7}$ & 27.2$_{\pm 1.0}$ & 65.3$_{\pm 0.4}$ & 44.2$_{\pm 0.2}$ & 45.0$_{\pm 0.1}$ & 56.0$_{\pm 0.3}$ \\
    Rand+Nettack  & 24.3$_{\pm 0.3}$   & 32.1$_{\pm 1.1}$ & 30.5$_{\pm 1.4}$ & 22.4$_{\pm 1.2}$ & 63.3$_{\pm 0.5}$ & 31.2$_{\pm 0.9}$ & 46.7$_{\pm 0.6}$ & 53.9$_{\pm 1.2}$ \\
    Node+Nettack  & 27.4$_{\pm 0.4}$   & 37.2$_{\pm 1.3}$ & \underline{39.9$_{\pm 1.3}$} & 27.7$_{\pm 1.4}$ & 78.6$_{\pm 0.3}$ & 63.2$_{\pm 0.5}$ & 53.6$_{\pm 0.7}$ & 78.3$_{\pm 0.8}$ \\
    AFGSM         & 26.3$_{\pm 4.2}$   & 38.6$_{\pm 3.2}$ & 37.5$_{\pm 1.9}$ & 32.3$_{\pm 1.1}$ & 74.9$_{\pm 0.7}$ & 45.8$_{\pm 0.7}$ & 65.8$_{\pm 0.9}$ & 77.4$_{\pm 0.6}$ \\
    TDGIA         & \underline{29.5$_{\pm 2.8}$}   & \underline{44.2$_{\pm 2.2}$} & 39.4$_{\pm 1.1}$ & \underline{32.5$_{\pm 0.7}$} & 93.3$_{\pm 0.2}$ & 91.8$_{\pm 0.5}$ & 67.2$_{\pm 0.4}$ & \underline{84.2$_{\pm 1.1}$} \\
    G-NIA         & 24.3$_{\pm 2.5}$   & 36.5$_{\pm 3.1}$ & 34.4$_{\pm 1.4}$ & 25.2$_{\pm 1.4}$ & \underline{95.0$_{\pm 0.4}$} & \underline{94.6$_{\pm 1.2}$} & \underline{68.3$_{\pm 1.0}$} & 81.1$_{\pm 1.2}$ \\
    \midrule
    G$^2$A2C (ours)& \textbf{36.3$_{\pm 2.7}$} & \textbf{49.4$_{\pm 2.8}$} & \textbf{42.2$_{\pm 1.1}$} & \textbf{33.6$_{\pm 1.2}$} & \textbf{97.4$_{\pm 0.4}$} & \textbf{98.7$_{\pm 0.6}$} & \textbf{74.1$_{\pm 0.8}$} & \textbf{86.6$_{\pm 0.8}$} \\

    Avg. $\uparrow$ & 6.8 & 5.2 & 2.3 & 1.1 & 2.4 & 4.1 & 5.8 & 2.4 \\
    \bottomrule
    \bottomrule
  \end{tabular}
  % \vspace{-0.2cm}
  \caption{Miclassification rate (\%) of a trained two-layer GCN model after the single node injection attack (i.e., $\beta_{e}=1$, $\beta_{n}=1$, and $\beta_f=0$) launched by the different attackers. \textbf{Rate} in bold indicates the best and \underline{rate} in underline is the second best. The results reported above are averaged over 10 independent runs with different initialization seeds.}
  \label{tab:exp_overall}
  % \vspace{-0.5cm}
 \end{table*}

\subsection{Experimental Setup} 
For the baselines, we explore DeepRobust \cite{li2020deeprobust} and open-source code with the default settings. 
We set the hyper-parameters in G$^2$A2C as following: the number of GCLs $K$ to 2, the temperature of Gumbel-Softmax to 1.0, hidden dimension $d$ to 256, and the discount factor to 0.95. 
We utilize Adam optimizer with learning rate $10^{-4}$. Besides, we adopt the early stopping with a patience of 3 epochs. 
All experiments are conducted for 10 times with mean and deviation reported. 
% Software we use includes Torch 1.9.0, DGL 0.9.1 and Pyro 1.8.0. Experiments run on a server with AMD Ryzen 3990X CPU, RTX 2080ti, and 128GB RAM.

\subsection{Performance Comparison}
We perform single injection attack, the most extreme setting where only one injected node with one edge is allowed to attack a target node.
The results of G$^2$A2C and all baselines are reported in Table \ref{tab:exp_overall}. 
We can observe that all baselines cause a performance downgrade to the victim model, and the introduced perturbation on the datasets with continuous feature space is more severe than those with discrete feature space, demonstrating the obstacle from the unsmooth feedback during the attack phase. 
Compared with all baselines, on average, G$^2$A2C increases the misclassification rate by 3.85 on discrete datasets and by 3.68 on continuous datasets. 
We further conduct a t-test on the results of G$^2$A2C and the best performing baseline in each dataset, and the performance improvement brought by G$^2$A2C is significant with 95\% confidence, demonstrating the superior and stable performance of G$^2$A2C. 
By comparing perturbation models of random features with those utilizing the generated adversarial features from G$^2$A2C (e.g., Rand+Nettack vs. Node+Nettack), we can clearly observe that the latter greatly imperils the performance of the victim model, and all baselines can cause considerable downgrade with the generated features from G$^2$A2C, which further demonstrates the legitimacy of the node generator in G$^2$A2C. 
To answer \textbf{RQ1}: under the black-box setting, G$^2$A2C outperforms all baselines and by a significant margin across all datasets in the most extreme setting. 
Besides, the performance of weak baselines can be significantly boosted by the adversarial features generated by G$^2$A2C, as demonstrated by Node+NIPA and Node+Nettack. 
By comparing Node+Nettack with G$^2$A2C, we observe a higher performance for G$^2$A2C, indicating the outstanding edge wiring capability of the edge sampler. 
\begin{table}
\centering
 \begin{tabular}{l|c|cccc}
    \toprule
    Attacker & Backbone & Cora & Citeseer & Pubmed \\
    \midrule
    \multirow{4}{*}{G-NIA} & GCN & 24.3$_{\pm 2.5}$ & 36.5$_{\pm 3.1}$ & 68.3$_{\pm 1.0}$\\
    & SGC                        & 27.6$_{\pm 2.9}$ & 40.1$_{\pm 3.5}$ & 70.2$_{\pm 1.1}$ \\
    & GAT                        & 22.1$_{\pm 2.1}$ & 35.3$_{\pm 1.8}$ & 68.8$_{\pm 0.7}$ \\
    & APPNP                      & 24.2$_{\pm 2.0}$ & 37.2$_{\pm 3.4}$ & 69.7$_{\pm 0.8}$ \\
    \midrule
    \multirow{4}{*}{TDGIA} & GCN & 29.5$_{\pm 2.8}$ & 44.2$_{\pm 2.2}$ & 67.2$_{\pm 0.4}$ \\
    & SGC                        & 35.2$_{\pm 2.1}$ & 51.2$_{\pm 1.1}$ & 74.4$_{\pm 1.5}$ \\
    & GAT                        & 28.3$_{\pm 2.0}$ & 52.1$_{\pm 1.4}$ & 81.2$_{\pm 0.2}$ \\
    & APPNP                      & 29.1$_{\pm 2.4}$ & 43.8$_{\pm 1.1}$ & 69.6$_{\pm 2.9}$ \\
    \midrule
    \multirow{4}{*}{G$^2$A2C} & GCN & 36.3$_{\pm 2.7}$ & 49.4$_{\pm 2.8}$ & 74.1$_{\pm 0.8}$ \\
    & SGC                           & 41.6$_{\pm 3.1}$ & 53.8$_{\pm 1.4}$ & 76.3$_{\pm 2.2}$\\
    & GAT                           & 33.4$_{\pm 3.3}$ & 55.6$_{\pm 1.6}$ & 85.0$_{\pm 0.3}$ \\
    & APPNP                         & 36.4$_{\pm 2.7}$ & 48.0$_{\pm 1.3}$ & 82.9$_{\pm 3.2}$ \\
    \bottomrule
  \end{tabular}
  \caption{Misclassification rate (\%) to different two-layer GNNs. The same setting as reported in Table \ref{tab:exp_overall} is explored. }
  \label{tab:all_gnn}
  % \vspace{-0.5cm}
 \end{table}
 
To further validate the performance of G$^2$A2C, we change the backbone GNN of the victim model to GAT \cite{velivckovic2017graph}, APPNP \cite{klicpera2019predict}, and SGC \cite{wu2019simplifying} and launch attacks by the best-performing baselines as well as G$^2$A2C.
In this setting, we do not accordingly modify the surrogate model architecture or the backbone GNN of G$^2$A2C to intentionally create an extreme setting where the attacker possesses an inaccurate approximation of the victim model. 
The results are shown in Table \ref{tab:all_gnn}. 
We can observe that the performances of all attackers are significantly enhanced on shallow models such as SGC, indicating the vulnerabilities of the shallow and less-parametrized models.
While attacking the over-parametrized GAT model, G-NIA delivers downgraded performances, but TDGIA and G$^2$A2C show stronger performance compared with the performance of attacking other shallow models (e.g., GCN and SGC). 
This phenomenon demonstrates that the vulnerabilities of GAT is dependent on the different datasets.
Overall, G$^2$A2C outperforms all best-performing baselines over these three exemplary datasets. To answer \textbf{RQ2}, when the attacker's assumptions on the architecture of the victim model are incorrect, the effectiveness of G$^2$A2C is barely downgraded, demonstrating the advantages brought by the ``gradient-free'' property.

\subsection{Budget Analysis}
\begin{figure}
\centering
% \vspace{-0.5cm}
\includegraphics[width=1.\linewidth]{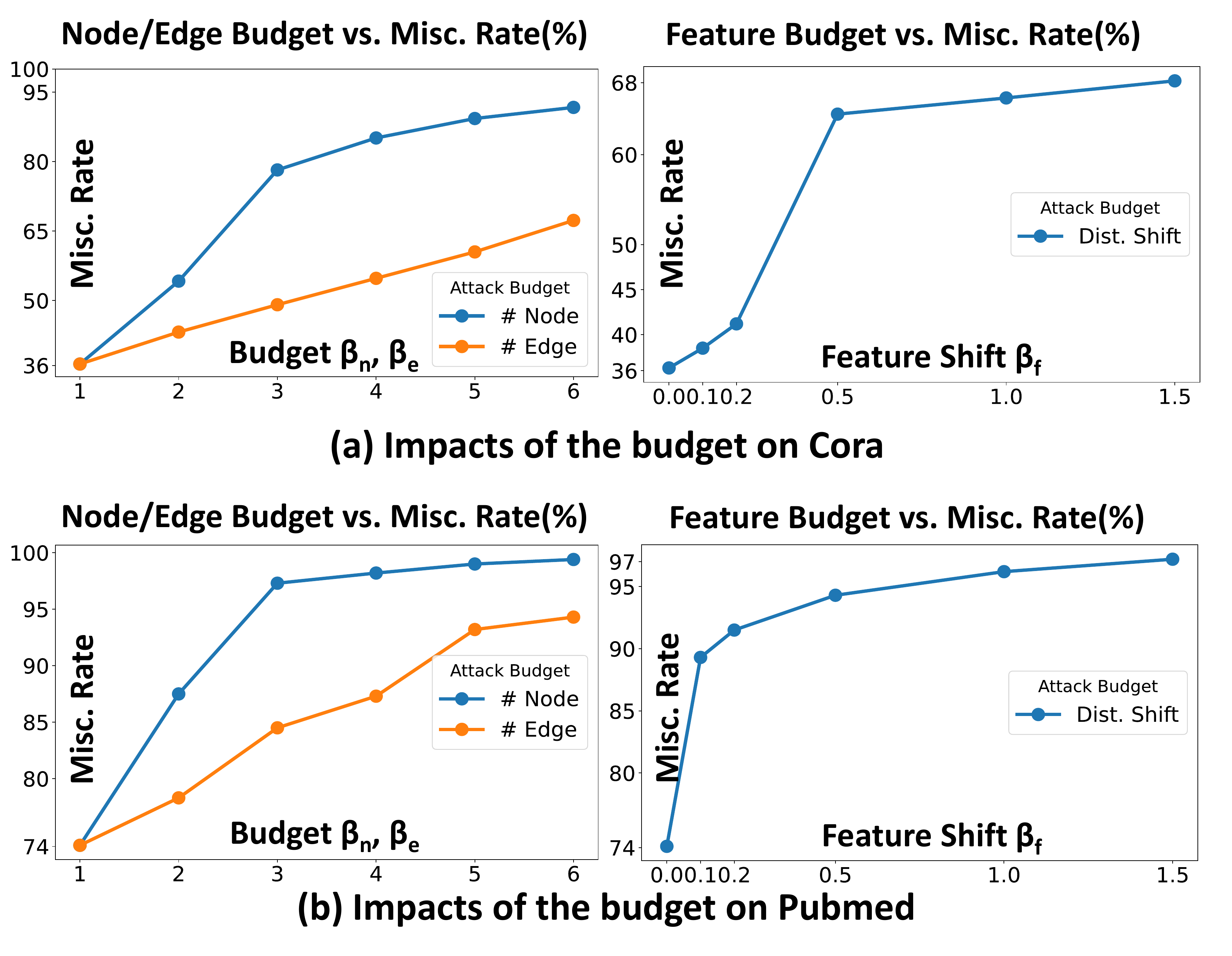}
% \vspace{-0.7cm}
\caption{Effectiveness of G$^2$A2C under different budgets.}
% \vspace{-0.3cm}
\label{fig:budget}
\end{figure}

In this section, we conduct experiments w.r.t. all the attack budgets in our setting: the edge budget $\beta_e$, the node budget $\beta_n$ and the feature shift budget $\beta_f$, as shown in Figure \ref{fig:budget}.
From these results, to answer \textbf{RQ3}, the most fruitful budget is $\beta_n$, and with 3 injected nodes per target, the performance of GCN on all datasets is downgraded to the range around 10\%. 
The second most impactful budget is $\beta_f$, and the performance of GCN all falls down below 10\% with a distribution shift budget of 0.5 and its impact saturates around 0.5. 
The least impactful budget is $\beta_e$. We observe that its impact is relatively linear compared with other budgets. 

\subsection{Case Study}
\begin{figure}
\centering
% \vspace{-0.5cm}
\includegraphics[width=1\linewidth]{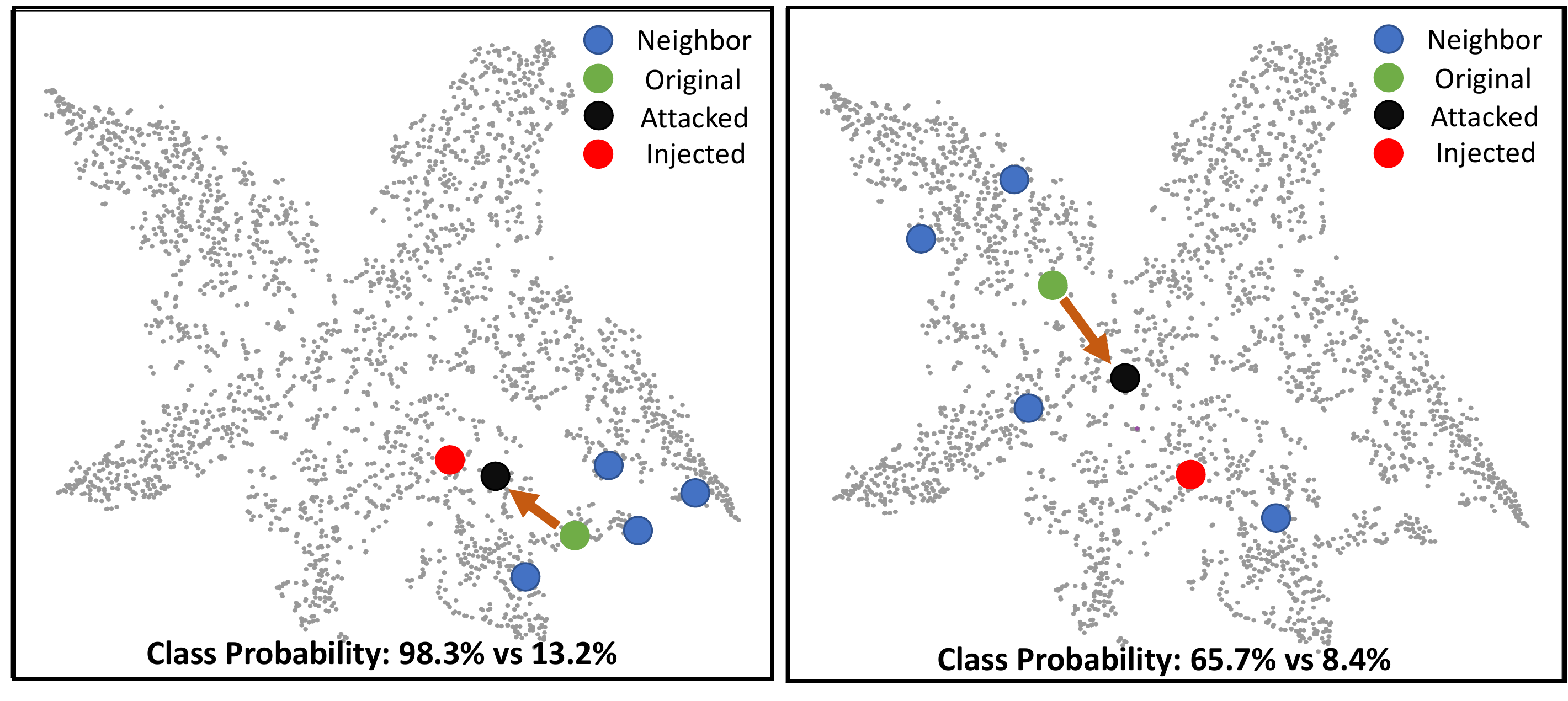}
% \vspace{-0.3cm}
\caption{Visualization of the attack launched by G$^2$A2C.}
\label{fig:tsne}
% \vspace{-0.2cm}
\end{figure}

To further investigate how G$^2$A2C conducts node injection attack, we visualize two successful attacks on Citeseer. 
As shown in Figure \ref{fig:tsne}, we visualize the attack process by plotting the hidden embedding of the involved nodes, extracted from the victim model, before and after the attack via T-SNE \cite{van2008visualizing}. 
In this figure, blue points are target node's original neighbors in the clean graph, green point is the target node before the attack, black point refers to the attacked target node, and red point refers to the injected adversarial node. 
To answer \textbf{RQ4}, as shown in these two cases, the injected node could effectively perturb the embedding of the target node, relocate it to a relatively intertwined position, and hence flip its prediction.

\section{Related Work}
GNNs have been proved to be sensitive to adversarial attacks \cite{dai2018adversarial,ma2019attacking,wang2019attacking,xu2019topology,zugner2018adversarial,sun2019node,tao2021single,wang2020scalable}. 
Most of them focus on perturbations on existing knowledge, such as topological structure \cite{xu2019topology,wang2019attacking,ma2019attacking}, node attributes \cite{zugner2018adversarial}, and labels \cite{sun2019node}. 
However, in the real world, modifying existing edges or node attributes is not practical, due to limited access to the node of interests. 
Node injection attack aims at a more realistic scenario, which adds adversarial nodes in to the existing graph. 
State-of-the-art attackers \cite{sun2019node,tao2021single,wang2020scalable, zou2021tdgia} either explore the less practical white-box setting, or training a surrogate model to simulate a white-box setting, which might introduce performance downgrade when inaccurate approximations about the victim model are made. 
For the white-box setting, NIPA \cite{sun2019node} creates a batch of random nodes and wiring them to the existing graph to fulfill the malicious intent. 
And for the black-box setting, AFGSM \cite{wang2020scalable} utilizes a fast gradient sign method and G-NIA \cite{tao2021single} explores a neural network to generalize the attacking process. 
TDGIA firstly selects topological defective edges to the injected node, and then generates the adversarial features for the injected nodes according to the surrogate model. 
To further leverage the practicality as well as effectiveness, we study the node injection attack under the black-box setting without training a surrogate model to acquire simulated gradient, eliminating the possibility of error propagation due to inaccurate approximations about the victim model.
\section{Conclusion}
In this work, we study gradient-free node injection evasion attack for graphs under the black-box setting. 
Unlike other node injectors requiring gradient from the surrogate model, we propose G$^2$A2C, a gradient-free attacker without any assumption on the victim model, eliminating the possibility of error propagation due to inaccurate approximations about the victim model.
We formulate such attack as an MDP and solve it through our designed graph reinforcement learning framework. 
Our node generator generates imperceptible yet malicious node features, followed by the edge sampler that wires the node to the remaining graph. Through comprehensive experiments with the state-of-the-art baselines, we demonstrate the promising performance of G$^2$A2C over eight acknowledged datasets with diverse characteristics.
And by modifying the architectures of the victim model to four different GNNs, we empirically prove the advantage brought by the ``gradient-free'' property of G$^2$A2C.

\section*{Acknowledgments}
This work is partially supported by the NSF under grants IIS-2209814, IIS-2203262, IIS-2214376, IIS-2217239, OAC-2218762, CNS-2203261, CNS-2122631, CMMI-2146076, and the NIJ 2018-75-CX-0032. Any opinions, findings, and conclusions or recommendations expressed in this material are those of the authors and do not necessarily reflect the views of any funding agencies.

\bibliography{aaai23}

\end{document}